\documentclass[numbers]{article}

\usepackage[preprint]{neurips_2023}

\PassOptionsToPackage{numbers}{natbib} 
\usepackage{natbib}  
\usepackage[utf8]{inputenc} 
\usepackage[T1]{fontenc}    

\usepackage{url}            
\usepackage{booktabs}       
\usepackage{amsfonts}       
\usepackage{nicefrac}       
\usepackage{microtype}      
\usepackage{xcolor}         

\usepackage{amsmath,amsfonts,amssymb}
\usepackage{graphicx}

\usepackage{algorithm}
\usepackage{algpseudocode} 
\title{Computationally efficient non-Intrusive pre-impact fall detection system}

\author{%
  Praveen Jesudhas \\
  Department of Computer Science \& Engineering\\
  Anna University, Guindy\\
  Chennai, India\\
  \texttt{pjesudhas@gmail.com} \\
  \And
  Raghuveera T\\
  Department of Computer Science \& Engineering\\
  Anna University, Guindy\\
  Chennai, India\\
  \texttt{raghuveera@annauniv.edu} \\
  \And
  Shiney Jeyaraj\\
  Shark AI Solutions\\
  126, Judges Colony\\
  Chennai, India\\
  \texttt{shiney@sharkaisolutions.com} \\
}

\begin{document}

\maketitle

\begin{abstract}
 Existing pre-impact fall detection systems have high accuracy, however they are either intrusive to the subject or require heavy computational resources for fall detection, resulting in prohibitive
deployment costs. These factors limit the global adoption of existing fall detection systems. In this work we present a Pre-impact fall detection system that is both non-intrusive and computationally
efficient at deployment. Our system utilizes video data of the locality available through cameras, thereby requiring no specialized equipment to be worn by the subject. Further, the fall detection system
utilizes minimal fall specific features and simplistic neural network models, designed to reduce the computational cost of
the system. A minimal set of fall specific features are derived from the skeletal data, post observing the relative position of human skeleton during fall. These features are shown to have
different distributions for Fall and non-fall scenarios proving their discriminative capability. A Long Short Term Memory (LSTM) based network is selected and the network architecture
and training parameters are designed after evaluation of performance on standard datasets. In the Pre-impact fall detection system the computation requirement is about 18 times lesser than existing modules with a comparable
accuracy of 88\%. Given the low computation requirements and higher accuracy levels, the proposed system is suitable for wider adoption in engineering systems related to industrial and residential safety.
\end{abstract}

\section{INTRODUCTION}
Falling is an event that can cause irreversible damage to human beings especially among elderly people if not identified and treated early. The detrimental effects of fall can increase with age due to loss of strength in legs, medication side effects, vision problems, and strength reduction in other tissues. Fall related injuries result in pain, disability, and sometimes leading to premature death. Falls also bring out psychological burden, economic pressures and even impact the caregiver’s quality of life [1].

According to the studies reported by the World Health Organization (WHO)  [1], falls represent the second leading cause of accidental deaths around the world, producing a particularly high morbidity among people aged 65 and older. For those aged over 80 residing in community settings, the percentage of those persons who experience at least one fall per year climbs to 50\% [2], with 40\% of them suffering recurrent falls [3]. In the USA, the annual number of fall related injuries is expected to reach 3.4 million in 2020 and 5.7 million by the year 2030 [4]. As it refers to the economic impact on the sustainability of national health systems, the global medical costs attributable to falls in 2015 totalled about \$50 billion([5],[6].

Given the nature of fall, its impact on the victim and the overall economy, it is crucial to detect fall (pre-impact fall) before hitting the ground when possible. Pre-impact fall detection systems [7] identify fall anytime between 0.2 to 0.8 seconds before the impact caused by fall. Generally detecting fall earlier from the time of impact helps to open airbags or other supporting systems that can minimize or prevent the detrimental effects of the fall. These systems are broadly classified under wearable sensors [8], ambient sensors [9] and vision based systems[10]. Wearable sensors include sensors that a person actually wears or holds in close proximity such as an accelerometer  or a Gyroscope. These data sources provide detailed features representing the movement of the person and they have been successfully utilized to detect fall both in pre-impact and post-impact scenarios. However, the downside of this approach is that every person needs to have the wearable sensor placed, which can add to the cost and might be intrusive to wear throughout. This limits the global adoption of wearable sensors based fall detection systems.

The Ambient sensors based fall detection system utilizes Radar [11], Sound, Wi-fi or similar metrics [12] which identify the change in a certain emitted quantity based on the occurrence of fall. These systems are valuable since a single system can monitor multiple people in a locality and are also non-intrusive to the subject monitored. However they still require the placement of specialized devices for the recognition of fall [12]. It is found that the ambient sensors based fall detection systems work only in post-impact fall detection scenarios [13]. In vision based fall detection systems [14], the video feeds got from CCTV devices are used as the base data. The advantage of this system is that it can be scaled across different locations with existing CCTV devices with no requirement for separate hardware [15]. Apart from their ease of scalability, they are less cumbersome for the person since they donot have to wear any intrusive sensors. At present these systems are utilized for post-impact fall detection but haven’t yet been tried on Pre-impact fall detection scenarios.

In the proposed work, minimal set of features specific to fall is derived from 2D skeleton to reduce complexity and subsequently, computationally efficient models are trained to detect fall. The specific contributions are as follows: 1) Detailed analysis of the process of fall, derivation of fall specific skeletal features, and comparison of their distributions for fall and non-fall scenarios. 2) Selection and design of Neural network architecture for pre-impact fall detection systeml. 3) Comparison of our work with existing systems based on Accuracy and complexity with the UP-Fall [16]. The rest of this work is organized in the following manner: Section 2 outlines the overall pre-impact fall detection system along with the scope of our work. Section 3 details the analysis of fall in detail along with the selection of relevant fall features. Section 4 elaborates on the procedure followed to develop the pre-impact fall detection model. The post-impact Fall detection module development is detailed in Section 5 along with their performance evaluation with existing systems. Section 6 concludes this work with a summary and future direction.

\begin{figure*}
	\centering
	\includegraphics[width=0.6\linewidth]{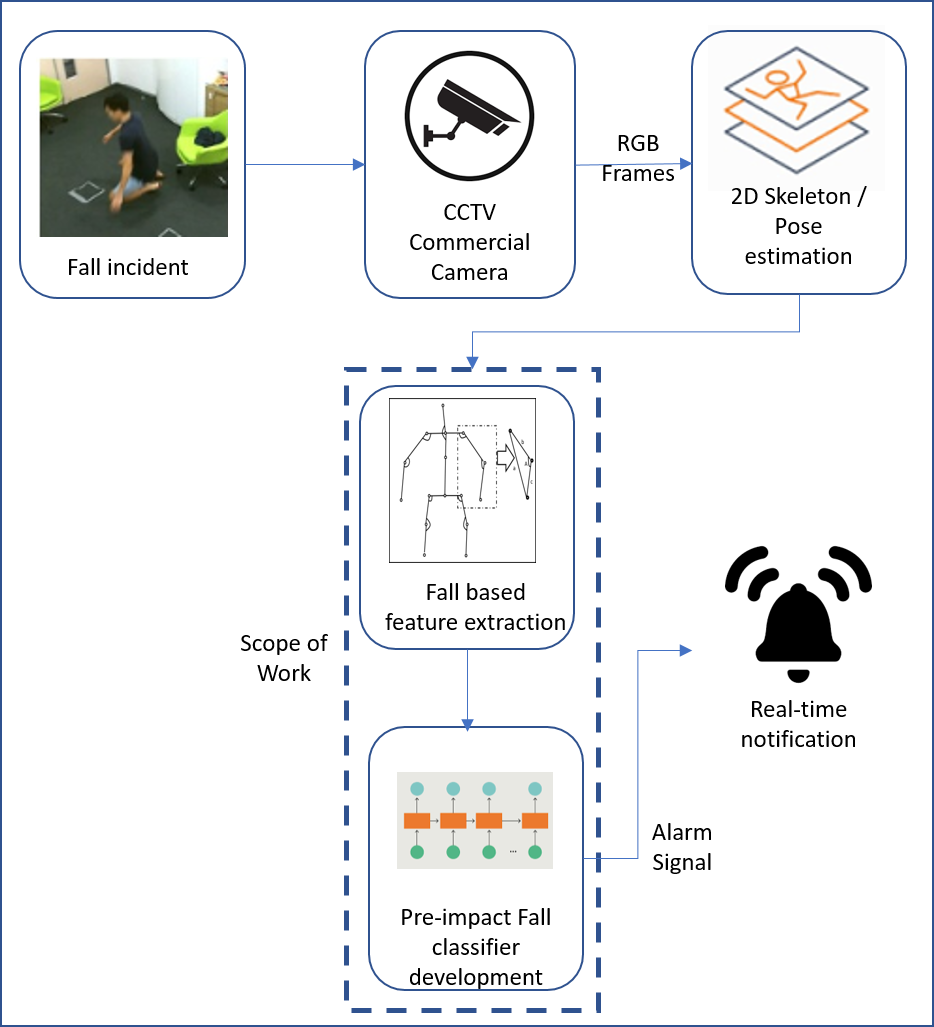}
	\caption{Overall System}
	\label{fig_overallsys}
\end{figure*}

\section{SCOPE OF WORK}\label{secrelated}
	In this section, the different modules involved in the pre-impact fall detection process are briefed and the significance of our work in the overall process is presented. The overall process of fall detection from CCTV camera videos is detailed in Fig.\ref{fig_overallsys}. Everyday activities performed by people in the field of view of the CCTV camera are captured in the form of RGB-based video streams. These streams are then converted to 2D skeletal data by Pose estimation algorithms such as Movenet [17]. These pose estimation modules are quite mature and are found to provide nearly 100\% accuracy in good quality video feeds. Fall as a process is analyzed in detail both from a spatial and temporal perspective and the regions of the human body impacted during fall is explored. The findings are correlated with regions in 2D skeletal data and representative skeletal features are identified that can help distinguish the occurrence of fall from ordinary day-to-day actions. The discriminative capability of these features is validated by the significant differences in their median values and distributions during fall and other actions

In the pre-impact fall detection system, these features are computed across definite intervals of time in different scenarios such as before the occurrence of fall and during other day-to-day actions. Subsequently, with the collated data, computationally efficient Neural net models based on the LSTM are designed post detailed experimentation. The results are validated across different types of fall in the UP-fall detection dataset [16]. The accuracy and complexity of the developed classifiers are compared with other state of the art models in fall detection. It is observed, that while maintaining similar levels of accuracy the system proposed in this paper had complexity at several orders of lower magnitude. This could help in the edge deployment of these models on low-cost infrastructure resulting in wider adoption in real-world environments.

\section{FALL-SPECIFIC FEATURE SELECTION}\label{secframe}
In this section, the process of fall is analyzed in detail and the minimal representative features that could identify a fall, before impact to the ground, are extracted. The distribution of these features over a fixed window size is analyzed between pre-impact fall and non-fall actions. Additionally, the same features are computed for both fall and non-fall over their entire duration separately, and their feature distributions are compared for their ability to distinguish post-impact fall from regular actions.

\subsection{Process of Fall}
\begin{figure*}
	\centering
	\includegraphics[width=0.95\linewidth]{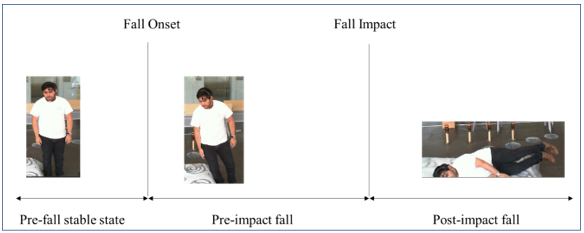}
	\caption{Different phases of Fall [16]}
	\label{fig2}
\end{figure*}

Falling in humans generally occurs due to a variety of reasons such as headaches, nature of surface, carelessness, drowsiness etc. It also occurs in different postures such as frontal, backward and along the sides. These conditions could in turn trigger imbalance in the lower limbs of the human body causing falls. For analysis, the activity of fall is segregated into three phases which are: ‘pre-fall stable state’, ‘pre-impact fall’ and ‘post-impact fall’. The different phases of fall are illustrated in Figure \ref{fig2}.

\begin{figure*}
	\centering
	\includegraphics[width=0.95\linewidth]{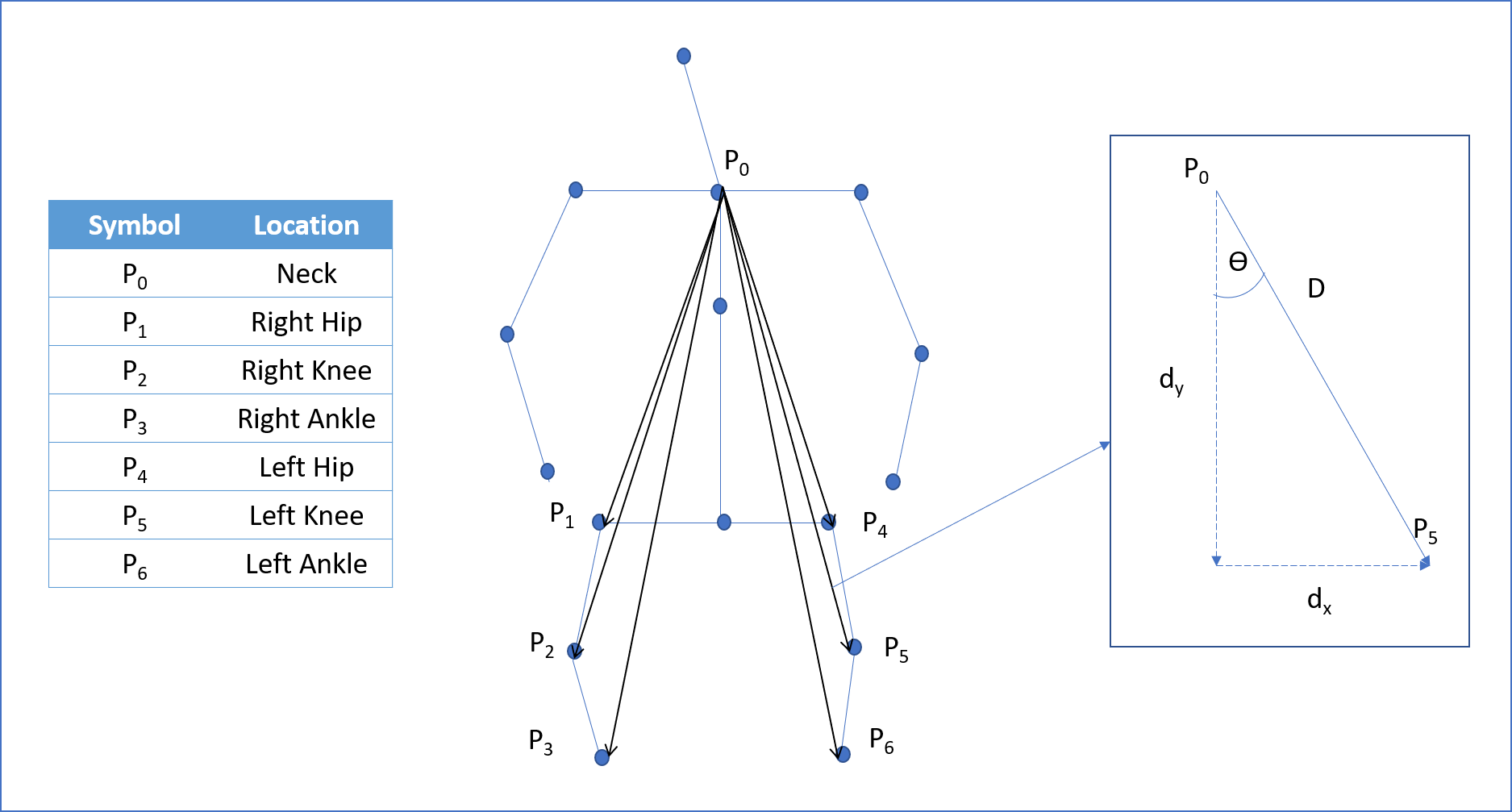}
	\caption{Fall Features representation}
	\label{fig_fallfeat}
\end{figure*}

The stable state before fall represents the action done by the person before the onset of fall. These could pertain to normal actions such as standing, sitting etc. At this state, there are no cues to identify fall and hence is out of interest for this work. Since the onset of fall, the lower body of people tends to start having imbalance, increasing significantly as the person hits the ground. This phase is called the pre-impact fall phase. The imbalance could be characterized by positional change at the lower limbs of the person of interest and they become more pronounced nearing fall impact. This insight is crucial to select high-sensitive skeletal features that can represent the minute variations in the lower limbs of human body to detect fall before impact as early as possible. Given this context, the activity of fall could be best represented by the variation in the position of joints across the lower part of the human body such as the ankle, knee and hip. High-sensitive fall features are derived from the 2D skeletal structure to identify the potential occurrence of fall before actual impact as illustrated in Fig. \ref{fig_fallfeat}. The description and computation of these features from the 2D skeletal data are detailed in the next subsection.


\subsection{Skeletal Fall features}
The 2D skeletal data extracted from pose estimation algorithms have the (x,y) camera coordinates of 25 different joints of the human body. These joints are consolidated across a period.

Considering the occurrence of fall affecting the relative position of the lower body joints in humans, the angles of the line segments connecting the different joints in the lower part of human body with the head are considered as features representing fall. The features are shown in Fig. \ref{fig_fallfeat}. The features are computed over a sequence of time to ensure the relative changes in these values as a fall occurs are captured for detecting falls. In subsequent sections, the high sensitive fall features are extracted from the 2D skeletons based on Pseudocode \ref{alg:cap} for both the pre-fall sequences for pre-impact fall detection and complete fall sequences for post-impact fall detection. 

\begin{algorithm}
\caption{Derivation of high sensitive Fall features }\label{alg:cap}
\begin{algorithmic}
\For {frame in a video sequence }
\State Read 2D (x, y) coordinates for the seven joints in Fig. 3 
\For {feature P1 to P6 in Fig. 3 do }
\State Compute Euclidean distance D between P0 and each feature point Px 
\State Compute the horizontal distance dx as shown in Fig. 3 
\State Compute high sensitive fall features using:  \(\theta = \sin^{-1}{\frac{d_x}{D}}\) (taking the right quadrant into account)
\State  Accumulate high sensitive fall features within each frame 
\EndFor
\State Accumulate high sensitive fall features across entire video sequence
\EndFor
\end{algorithmic}
\end{algorithm}

\subsection{Pre-impact fall detection – feature analysis}

\begin{figure*}
	\centering
	\includegraphics[width=0.6\linewidth]{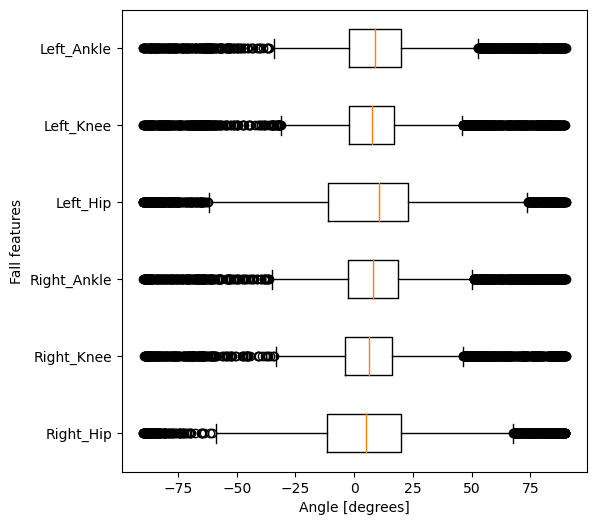}
	\caption{Feature distribution for Pre-impact fall}
	\label{fig_featpre}
\end{figure*}

\begin{figure*}
	\centering
	\includegraphics[width=0.6\linewidth]{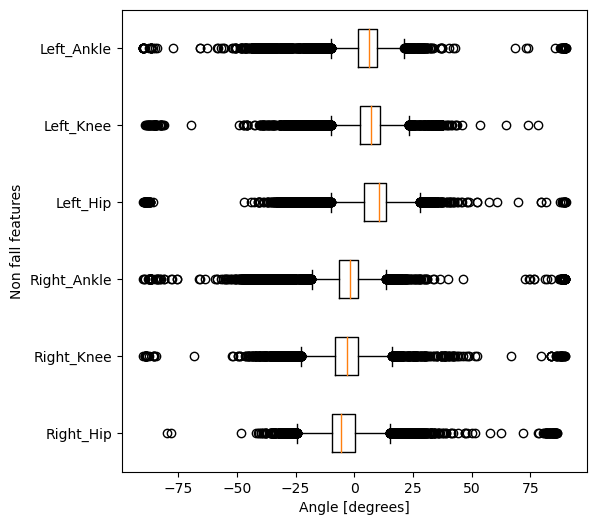}
	\caption{Feature distribution for non-fall (Pre-impact fall)}
	\label{fig_featno}
\end{figure*}

To analyse and compare the derived high sensitive fall features from the perspective of pre-impact fall detection, the UP-Fall detection dataset is utilized. This dataset has 17 different subjects performing 11 different activities in 3 trials. Among the 11 activities, 5 activities represent different types of falls and the remaining 6 are non-fall actions. For analysis, 20 sequential frame samples are selected for all the fall sequences at 0.5 seconds lead time, before the impact to the ground. Similarly, for the non-fall sequences, 20 sequential frame samples are selected at random intervals over the course of the action. The proposed features are computed on the contiguous frame samples selected and grouped separately for the fall and non-fall related actions. The distribution for each of the high sensitive fall features is visualized in the form of box plots shown in Figures \ref{fig_featpre} and \ref{fig_featno}. The distribution of the features for Pre-impact fall, shows that the highest inter-quartile range is observed for hips, where it ranges from -15 to 20 degrees. It is also observed that the min and max values for the hips vary from -60 to 60 degrees.  On comparison with the same features for non-fall actions, it could be observed that the maximum interquartile ranges for the hip have a range between -8 to 0 degrees and the min-max values have a range between -25 to 15 degrees. 

It could be inferred from the above observations, that the variance of skeletal features observed for fall scenarios is significantly higher than for non-fall scenarios. This difference in the distribution of the skeletal features between fall and non-fall scenarios brings better discriminative ability in distinguishing pre-impact fall from non-fall scenarios. Hence these high-sensitive features could be considered to develop models for detecting pre-impact fall.

In the next sections, we focus on developing Neural net based models that can detect fall in pre-impact scenarios from the features derived in this section.

\section{PRE IMPACT FALL DETECTION MODULE DEVELOPMENT}\label{secframe}
This section starts with an overall view of the pre-impact fall detection system during its training and inference phases. The neural net architecture is elaborated in detail followed by model training, hyperparameter tuning and validation.

\begin{figure*}
	\centering
	\includegraphics[width=0.8\linewidth]{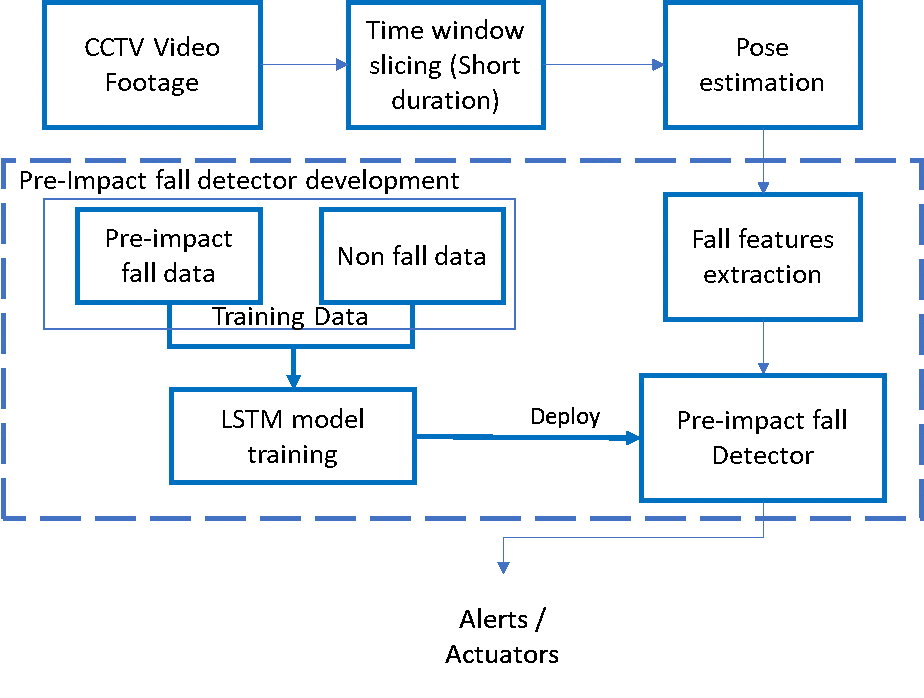}
	\caption{Pre-impact Fall detection - Overall system}
	\label{fig_preall}
\end{figure*}  
\subsection{System Overview}
The proposed system is elaborated in \ref{fig_preall}. The neural network model is trained with the high-sensitive skeletal features described in section 3.2, and is extracted from both the fall and non-fall video sequences. At inference, video streams from the CCTV footage are divided into image sequences that have shorter durations. The pose information is extracted from these images at a frame level, and the fall features are subsequently computed. The fall features are passed as input to the previously trained pre-impact fall detection model. The occurrence of a fall is determined from the output of the trained neural net model and is subsequently used to trigger necessary alerts. The overall system is illustrated in Figure \ref{fig_preall}.

The neural network model responsible for detecting fall, needs to utilize the features discussed in section 3.2 across the duration of interest. Apart from utilizing the information encoded by features at every frame they also need to learn the variation signature of these features across time. Given this context, sequence-based neural network frameworks are best suited for the fall detection use case. Among sequence based neural networks, LSTM framework is preferred for the model over a GRU and vanilla RNN networks due to its ability to capture both long and short-term dependencies coupled with no vanishing gradient issues.

\begin{figure*}
	\centering
	\includegraphics[width=0.5\linewidth]{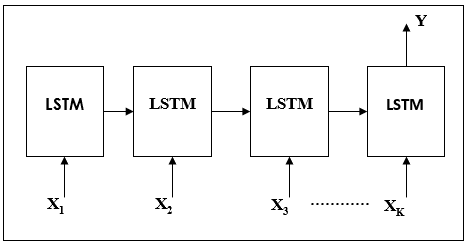}
	\caption{LSTM Architecture - Pre-Impact Fall detection}
	\label{fig_stagefalldetect}
\end{figure*}  
\begin{table}
	\caption{UP-Fall detection Dataset}
           \centering
	\begin{tabular}{p{7cm}p{2cm}}
		\hline
		Description& Data Size\\
		\hline
		No. of Subjects&	17\\
		No. of activities&	11\\
		No. of Fall activities&	5\\
		No. of Non Fall activities&	6\\
		No. of trials&	 3\\
		No. of cameras&	2\\
		\hline
	\end{tabular}
	\label{table1}
\end{table}
\subsection{Fall Detection Model Framework}

The primary requirement from a model development standpoint is computational efficiency while maintaining high accuracy standards. In an LSTM based framework, computational efficiency could be achieved by reducing the number of weights/parameters present in the LSTM network. The number of weights/parameters is directly proportional to the learning ability which necessitates striking a balance between the number of weights in the network and the accuracy of the system.

Different variations of the network size are experimented with model training parameters and the test-train accuracy curves are visualized. The optimal structure and weights of the LSTM based network are finalized based on the smallest network yielding higher accuracy levels on both the training and testing phases. The different set of parameters influencing the LSTM structure are Unidirectional/Bidirectional, no. of layers in the network and size of the hidden units in each LSTM cell. In terms of training the network, the loss function, optimizer type and batch size are important to train an accurate model. 

A unidirectional LSTM is selected since the input features representing fall are causal. Single layer network is selected to reduce the computational requirements of the network. The LSTM network being a binary detection system a negative log-likelihood loss function is selected. Adam optimizer is utilized for training due to its better convergence abilities. The structure of the LSTM network with these specifications are shown in \ref{fig_stagefalldetect} 

The features \(X_1\) to \(X_k\) are the high-sensitive fall features computed from each contiguous frame in the training dataset across a duration of time. The dimension of X at each frame is (1,6). They are then passed as input to their respective LSTM cells connected sequentially. The output of each LSTM cell is a function of the past LSTM cell outputs and the feature input at the current instant. The number of LSTM cells (K) connected sequentially is equal to the number of samples/frames in time window.
The output of the final LSTM cell is used as the classifier output Y. Y is a vector of size (1, 2) where each value represents the probability of fall vs non-fall. The index of Y which has the maximum value represents the predicted class of the model. 

\subsection{Model training and parameter selection}

To identify the optimal number of hidden units and batch size, the LSTM model is trained and tested with different configurations and the optimal parameters are selected by observing their test-train accuracy curves. The UP-fall detection dataset tabulated in Table 1, is the base dataset for training the pre-impact fall detection model.

The samples corresponding to the fall actions and non-fall actions are grouped separately. In the samples within Fall action, as observed in \ref{fig2}, there are three stages namely, the pre-fall stable state, pre-impact fall and post-impact fall. For experimentation purposes, the video frames are considered about 0.5 seconds before fall impact with a window size of 15 samples. Similarly, for the non-fall scenarios, a random start location in the frame sequence is selected and 15 continuous frames are grouped. High-sensitive fall features are derived from these video sequences as mentioned in Pseudocode 1 and a data set is collated. The dataset is split in the ratio of 80:20 for train and test purposes.

LSTM models with different configurations of hidden units and batch size are trained with the above dataset and test-train accuracy curves are observed. The model configurations that provide high accuracy with both the train and test data with minimal number of hidden units are selected as the optimal network for the pre-impact fall detection system. The accuracy is found to improve as the no of LSTM hidden units increases from 2 to 5 to provide a test accuracy F1 score of about 88\% post which the accuracy drops due to overfitting. A batch size of 8 is found to work well for this problem where the learning is stable. Considering these observations, a batch size of 8 and an LSTM hidden unit size of 5 units were selected for the final pre-impact fall detection model. The accuracy report for the final pre-impact fall detection model is shown in Table 2.

\begin{table}
	\caption{Pre-Impact Fall Detection Accuracy Report}
          \centering
	\begin{tabular}{p{4cm}p{2cm}p{2cm}p{2cm}}
		\hline
		 &Precision&	Recall&	F1 - Score\\
		\hline
		Pre-Impact Fall&	0.93&	0.72&	0.81\\
		Non-Fall&	0.9&	0.98&	0.94\\ 	 	 	 
		Macro Average&	0.92&	0.85&	0.88\\
		Weighted average&	0.92&	0.92&	0.92\\
		\hline
	\end{tabular}
	\label{table2}
\end{table}

\subsection{Analysis across Lead time and window sizes}

In this subsection, the accuracy of the pre-impact fall detection model at different lead time before impact is explored along with the optimal window size for the pre-impact fall detection model.

\begin{figure*}
	\centering
	\includegraphics[width=0.7\linewidth]{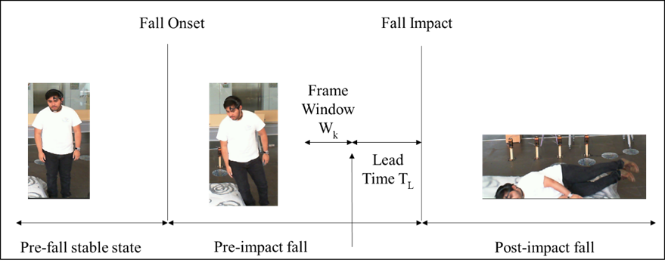}
	\caption{Stages of Fall Detection [16]}
	\label{fig_postall}
\end{figure*}

For evaluation purposes, contiguous frames within a window \(W_k\) are selected at a certain lead time before impact \(T_L\) as shown in \ref{fig_postall}. Different combinations of the training data with window size k ranging between 5 to 20 frames and lead time before fall ranging from 0.1 to 0.9 seconds have been considered. For the non-fall scenarios, a random start location in the frame sequence is selected.
To observe the model's accuracy across different window sizes, contiguous frames with varying window sizes (5, 10, 15, 20) are trained and validated. For the model training and validation, datasets in fall and non-fall scenarios having similar window sizes are collated. On these grouped datasets of frame sequences, the 2D skeletal information is extracted using pose estimation modules such as Movenet. The high-sensitive fall features mentioned in section 3.2 are then extracted based on Pseudocode 1. The data at each lead time and window size is split into train and test datasets for analysis.

The LSTM model is trained with high sensitive fall features for different variants of lead time and window sizes, and their accuracy is validated on the test dataset. The F1-scores of the trained model with lead time ranging from 0.9 to 0.1 seconds for a constant window size of 15 are shown in Table 3. Similarly in Table 4, the F1-scores of the models for different window sizes is tabulated for a constant lead time of 0.5 seconds. 
\begin{table}
	\caption{Pre-Impact fall detection (Impact time variation)}
           \centering
	\begin{tabular}{p{1cm}p{2.5cm}p{2cm}p{2.5cm}}
		\hline
		S. No&	Time before impact (seconds)&	Window Size&	Macro Average F1 Score\\
		\hline
		1&	0.1&	15&	0.93\\
		2&	0.2&	15&	0.93\\
		3&	0.3&	15&	0.91\\
		4&	0.4&	15&	0.9\\
		5&	0.5&	15&	0.88\\
		6&	0.6&	15&	0.86\\
		7&	0.7&	15&	0.74\\
		8&	0.8&	15&	0.69\\
		\hline
	\end{tabular}
	\label{table3}
\end{table}
\begin{table}
	\caption{Pre-Impact fall detection (window size variation))}
           \centering
	\begin{tabular}{p{1cm}p{2.5cm}p{2cm}p{2.5cm}}
		\hline
		S. No&	Time before impact (seconds)&	Window Size&	Macro Average F1 Score\\
		\hline
		1&	0.5&	5&	0.86\\
		2&	0.5&	10&	0.87\\
		3&	0.5&	15&	0.88\\
		4&	0.5&	20&	0.88\\
		\hline
	\end{tabular}
	\label{table4}
\end{table}
It could be observed from Table 3 that the accuracy increases with a decrease in lead time before the fall impact due to the availability of better signals representing the fall. The results in Table 4, show that the accuracy improves marginally when the number of frames in the window is increased from 5 to 15 frames, post which there is no improvement in accuracy. Though the highest accuracy is observed at a lead time of 0.1 seconds, the time interval needed to take corrective action is very minimal. Based on the reflexes of the human nervous system, it takes at least 0.5 seconds for a person to receive and effectively respond to external stimuli. For this purpose, the model developed at a lead time of 0.5 seconds could be considered for practical deployment in identifying pre-impact fall. The window size of 15 could be selected since the accuracy saturates post this size. 

\section{RESULTS AND DISCUSSION}\label{secframe}

In this section, the results of pre-impact fall detection models developed in section 4 and 5 are compared with other relevant work in terms of their complexity and accuracy. 
 
\subsection{Pre-impact fall detection – Performance comparison}
The pre-impact fall detection module developed is compared with other approaches both in terms of accuracy and complexity. It should be noted that the other approaches utilize sensor information such as accelerometers whereas our module utilizes only the 2D skeletal data acquired from CCTV cameras.

The other approaches are validated on the K-Fall dataset that contains the sensor-level information for each fall. The video data is not available in K-Fall dataset, due to which we have compared our finding on a similar UP-fall dataset which have the video feeds made public.
It could be observed that our pre-impact fall detection module uses much fewer parameters compared to existing approaches as shown in Table 5. The parameter space is found to be lesser by a factor of up to 80 times compared to existing approaches while having a slight drop in the accuracy when compared with our model having lead time of 500 ms. Our approach at 200 ms lead time is found to provide comparable levels of accuracy with the sensor-based methods to detect pre-impact fall.

Thus, it could be observed our approach utilizing 2D skeletal information from CCTV cameras is able to provide comparable accuracy to existing systems at a much lesser system complexity. The less complex nature of the system can help in much faster inference to identify fall before their occurrence using minimal edge hardware. The ability to detect Pre-impact fall from CCTV cameras makes our approach much more scalable for practical usage compared to sensor-based modules.                                                                      

\begin{table}
	\caption{Pre-Impact Fall Detection Performance Comparison}
	\centering
	\begin{tabular}{p{3cm}p{1.5cm}p{1.5cm}p{1.5cm}p{1.5cm}p{1.5cm}}
		\hline
		Model name&	Lead time (ms)&	Parameters (Byte)&	Precision&	Recall&	F1-score\\
		\hline
		Baseline-CNN [18]&	511.2±79&	59557&	92.36&	80.27&	85.89\\
		Lightweight-CNN[18]&	537.2±71&	59557	&84.97&	96.58&	90.4\\
		CNNLSTM [18]&	493.5±62&	81093&	88.35&	94.58&	91.36\\
		ViT-tiny [18]&	235.4±62&	251010&	92.02&	95.73&	93.84\\
		PreFallKD [18]&	551.3±66&	59557	&90.62&	94.79&	92.66\\
		2D Fall feature (Proposed)&	500&	3136&	92.2&	85.4&	88.2\\
		2D Fall feature (Proposed)&	200&	3136&	94.2&	92.1&	93.13\\
		\hline
	\end{tabular}
	\label{table5}
\end{table}

\section{CONCLUSION AND FUTURE WORK}\label{secframe}

A computationally efficient, non-intrusive pre-impact fall detection system is developed to identify the occurrence of falls, utilizing 2D skeletal data extracted from commodity-level CCTV camera video streams. The 2D skeletal data is analysed in detail to detect pre-impact falls, and high-sensitive skeletal features are extracted. The utility of the high-sensitive skeletal features for pre and post-impact fall detection is demonstrated with sufficient level of exploratory data analysis. LSTM-based sequence models are built to detect pre-impact fall utilizing the derived features on the UP-Fall detection dataset. Different window sizes and lead times before fall are experimented. An F1-score of 88\% is achieved for the pre-impact fall detection at 0.5 seconds lead time. The Pre-impact fall detection models are found to utilize far fewer parameters compared to other existing approaches resulting in lower cost of deployment. This is also the first work to this date, where pre-impact fall detection is attempted from CCTV imagery. The ability to detect fall before impact from CCTV imagery brings huge advantages in the ease of adoption.

As part of future work, it is planned to validate the efficacy of the proposed fall detection system in natural settings. Handling aspects such as partial occlusion are planned to be explored in more detail. Additionally for pre-impact fall detection, it is planned to utilize multiple camera views to enhance the accuracy and to detect falling with longer lead times before impact. The different ways of notification before fall impact are aimed to be studied in detail to ensure timely attention is provided. From an adoption standpoint, remote deployment architectures are planned to be explored to facilitate deployment on edge systems.

\section*{References}

{
\small

[1] Bhasin, S., Gill, T. M., Reuben, D. B., et al. (2020). A randomized trial of a multifactorial strategy to prevent serious fall injuries. *New England Journal of Medicine*, 383(2), 129–140.

[2] Qiu, H., \& Xiong, S. (2015). Center-of-pressure based postural sway measures: Reliability and ability to distinguish between age, fear of falling and fall history. *International Journal of Industrial Ergonomics*, 47, 37–44.

[3] Beauchet, O., Sekhon, H., Schott, A.-M., et al. (2019). Motoric cognitive risk syndrome and risk for falls, their recurrence, and postfall fractures. *Journal of the American Medical Directors Association*, 20(10), 1268–1273.

[4] Ambrose, A. F., Paul, G., \& Hausdorff, J. M. (2013). Risk factors for falls among older adults: A review of the literature. *Maturitas*, 75(1), 51–61.

[5] Faes, M. C., Reelick, M. F., Joosten-Weyn Banningh, L. W., et al. (2010). Qualitative study on the impact of falling in frail older persons and family caregivers. *Aging \& Mental Health*, 14(7), 834–842.

[6] LeLaurin, J. H., \& Shorr, R. I. (2019). Preventing falls in hospitalized patients: State of the science. *Clinics in Geriatric Medicine*, 35(2), 273–283.

[7] Wu, Y., Su, Y., Feng, R., Yu, N., \& Zang, X. (2019). Wearable-sensor-based pre-impact fall detection system with a hierarchical classifier. *Measurement*, 140, 283–292.

[8] Ren, L., \& Peng, Y. (2019). Research of fall detection and fall prevention technologies: A systematic review. *IEEE Access*, 7, 77702–77722.

[9] Chhetri, S., Alsadoon, A., Al-Dala'in, T., et al. (2021). Deep learning for vision-based fall detection system: Enhanced optical dynamic flow. *Computational Intelligence*, 37(1), 578–595.

[10] Patel, A. N., Murugan, R., Maddikunta, P. K. R., et al. (2024). AI-powered trustable and explainable fall detection system using transfer learning. *Image and Vision Computing*, 149, 105164.

[11] Nguyen, T.-T., Cho, M.-C., \& Lee, T.-S. (2009). Automatic fall detection using wearable biomedical signal measurement terminal. In *Proc. IEEE EMBC 2009*, 5203–5206.

[12] Sadreazami, H., Bolic, M., \& Rajan, S. (2019). Fall detection using standoff radar-based sensing and deep CNN. *IEEE Trans. Circuits Syst. II*, 67(1), 197–201.

[13] Jesudhas, P., \& Tripuraribhatla, R. (2023). Low complexity fall detection using 2D skeletal data. In *Proc. IEEE M2VIP 2023*, 1–6.

[14] Jesudhas, P., \& Raghuveera, T. (2023). A novel computationally efficient approach to identify visually interpretable medical conditions from 2D skeletal data. *Computer Systems Science \& Engineering*, 46(3).

[15] Umer, M., Alarfaj, A. A., Alabdulqader, E. A., et al. (2024). Enhancing fall prediction in the elderly people using LBP features and transfer learning model. *Image and Vision Computing*, 145, 104992.

[16] Martínez-Villasenor, L., Ponce, H., Brieva, J., et al. (2019). UP-fall detection dataset: A multimodal approach. *Sensors*, 19(9), 1988.

[17] Bajpai, R., \& Joshi, D. (2021). Movenet: A deep neural network for joint profile prediction across variable walking speeds and slopes. *IEEE Trans. Instrum. Meas.*, 70, 1–11.

[18] Hewamalage, H., Bergmeir, C., \& Bandara, K. (2021). Recurrent neural networks for time series forecasting: Current status and future directions. *International Journal of Forecasting*, 37(1), 388–427.

}

\end{document}